# ChitroJera: A Regionally Relevant Visual Question Answering Dataset for Bangla

**Deeparghya Dutta Barua[1]*, Md Sakib Ul Rahman Sourove[1]*, Md Fahim[1,2]*, Fabiha Haider[1], Fariha Tanjim Shifat[1], Md Tasmim Rahman Adib[1], Anam Borhan Uddin[1], Md Farhan Ishmam[1,3] , Md Farhad Alam[1]**

[1]Research and Development, Penta Global Limited
[2]CCDS Lab, Independent University, Bangladesh
[3]Kahlert School of Computing, University of Utah
{deeparghya.csedu, farhan.ishmam, pdcsedu}@gmail.com

## Abstract

Visual Question Answer (VQA) poses the problem of answering a natural language question about a visual context. Bangla, despite being a widely spoken language, is considered low-resource in the realm of VQA due to the lack of proper benchmarks, challenging models known to be performant in other languages. Furthermore, existing Bangla VQA datasets offer little regional relevance and are largely adapted from their foreign counterparts. To address these challenges, we introduce a large-scale Bangla VQA dataset, ChitroJera, totaling over 15k samples from diverse and locally relevant data sources. We assess the performance of text encoders, image encoders, multimodal models, and our novel dual-encoder models. The experiments reveal that the pre-trained dual-encoders outperform other models of their scale. We also evaluate the performance of current large vision language models (LVLMs) using prompt-based techniques, achieving the overall best performance. Given the underdeveloped state of existing datasets, we envision ChitroJera expanding the scope of Vision-Language tasks in Bangla.

## 1 Introduction

Visual Question Answering (VQA), has gained relevance lately with the onset of transformer-based models facilitating a better understanding of language and context in different modalities. This has led to the focus in VQA research shifting from the perception of language and vision to understanding the reasoning of these opaque systems (Schwenk et al., 2022). VQA systems are also being used to aid in visual impairment (Gurari et al., 2018), enhance robotic systems (Sermanet et al., 2023), expedite the screening of medical conditions from relevant imagery (Lin et al., 2023), and so on. The performance of these systems is strongly coupled with the quality of the dataset that they are trained on (Gong et al., 2023; Quionero-Candela et al., 2009). With that in mind, VQA datasets in English are being designed with increasing complexity, leaning more towards advanced reasoning instead of simple answers (Marino et al., 2019; Schwenk et al., 2022). The landscape for most low-resource languages, however, speaks quite differently with the major pain point being the lack of substantial datasets that are able to address even the most basic of answers.

Despite being a language with around 278 million[1] speakers, Bangla has received very limited exposure to the domain of visual question answering. The issues are multifaceted, with one aspect being the lack of pre-trained vision language models (VLMs) to entertain the idiosyncrasies of the language, and the other being the lack of datasets tailored for this particular purpose, further compounding the issue of VLM unavailability.

Presently, only four instances of VQA datasets have been compiled for Bangla (Islam et al., 2022; Rafi et al., 2022; Romero et al., 2024), all of which have certain limitations as per Tab. 1. The first work presents two datasets, Bengali-VQA-v1 and Bengali CLEVR, built on top of the existing English VQA datasets — VQA v1 (Antol et al., 2015) and CLEVR (Johnson et al., 2017) respectively. The second work presents Bengali-VQA-2.0, compiled from the English VQA v2.0 dataset. Among these, Bengali-VQA-v1 and Bengali-VQA-2.0 offer limited applicability to non-trivial VQA tasks, as the questions are restricted to binary answers. The answers in Bengali CLEVR do encompass multiple classes, but the issue that plagues all these datasets is that the source images and texts are derived from English datasets, which lack the cultural and geographical context associated with Bangla. Also, Bengali VQA 2.0 magnifies its scale by relying on tactics that yield repetitive samples with minor differences. Lastly, CVQA (Romero et al., 2024) addresses some of these issues but as it explores 26 languages, the focus on Bangla is quite thin, yielding only 286 samples.

*Equal Contribution

[1]https://www.ethnologue.com/insights/ethnologue200/

| Datasets | #Q | #A | #I | Img Src. | Annot. | Val. | QT | CMA |
|---|---|---|---|---|---|---|---|---|
| Bengali VQA v1 | 5000 | 2 | 500 | English | gTrans | Authors | Binary | ✗ |
| Bengali CLEVR | 12291 | 1600 | 1271 | English | gTrans | Authors | WH | ✗ |
| Bengali VQA 2.0 | 13046 | 2 | 3280 | English | Manual | N/A | Binary | ✗ |
| CVQA (Bengali subset) | 286 | 780 | 136 | Bangla | Manual | Natives | WH | ✓ |
| ChitroJera (Ours) | 15292 | 5542 | 15147 | Bangla | GPT-4 Turbo | Experts | WH | ✓ |

Table 1: **Comparison between existing datasets** based on the number of questions (#Q), answers (#A), and images (#I), source of images (Img Src.), annotation (Annot.) and Validation (Val.) methods, Question Type (QT), and categorical metadata availability (CMA). gTrans means Google Translate.

With the intent of addressing the shortcomings within the existing space of Bangla VQA, our contributions are as follows:

**ChitroJera Dataset**: We propose a new dataset for Bangla VQA, named ChitroJera, comprising 15k images and questions, synthesized using OpenAI GPT-4 Turbo with curated prompting and validated by linguistic experts. The images and text have a Bangla geocultural flavor, *i.e.* that they capture the cultural connotations associated with the Bangla-speaking region. We ensure diversity by imposing restrictions on the number of questions per image. For better analysis, we provide a categorical breakdown of the samples based on the subject of the questions. Comparison between our dataset and the existing ones can be found in Table 1.

**Dual Encoder Models**: Due to the lack of VLMs that are aligned on Bangla text with regional images, we introduce novel dual encoder-based models that outperform existing unimodal models and VLMs trained on English data sources, showing promising performance at its scale.

**Extensive Experimentation**: We conduct experiments on our dataset using state-of-the-art and widely adopted text encoders, visual encoders, and multimodal models fine-tuned for this task. We observe the performance of dual encoder models with and without pretraining, with ablations on different pretraining objectives and batch sizes. Finally, we assess large language models via zero-shot prompting and achieve the best performance.

## 2 ChitroJera Dataset

The creation of the ChitroJera dataset involves a meticulous sourcing and annotation process intended to address Bangla VQA tasks while preserving cultural nuances. We achieve this by focusing on internet sources representative of the Bangla-speaking demographic.

### 2.1 Dataset Collection

The question-answer pairs in the dataset have been collected from existing Bangla images and image captions found in the BanglaLekhaImageCaptions (Rahman et al., 2019), Bornon (Shah et al., 2022), and BNATURE (Faruk et al., 2020) datasets. These sources contain images from the internet which have been collected using keywords relevant to Bangladesh and its vicinity, transitively ensuring that our dataset is regionally relevant. A more detailed breakdown of the distribution of these sources can be found in Table 2.

### 2.2 Data Preprocessing

The caption-image mismatches in some of the source datasets have been corrected manually, the images have been deduplicated, and the erroneous ICC color profiles have been removed. The same image may have multiple captions, and for images with more than 3 captions, we choose the longest and the shortest captions, and a third caption that has the highest BERTScore with the former two. The reason behind this choice is to capture a diverse depiction of the image. While shorter captions typically convey the broader context, longer captions tend to include finer details. These three (or less) captions are concatenated to form the textual ground truth context for the question generation. The caption to context generation process has been highlighted in Algorithm 1.

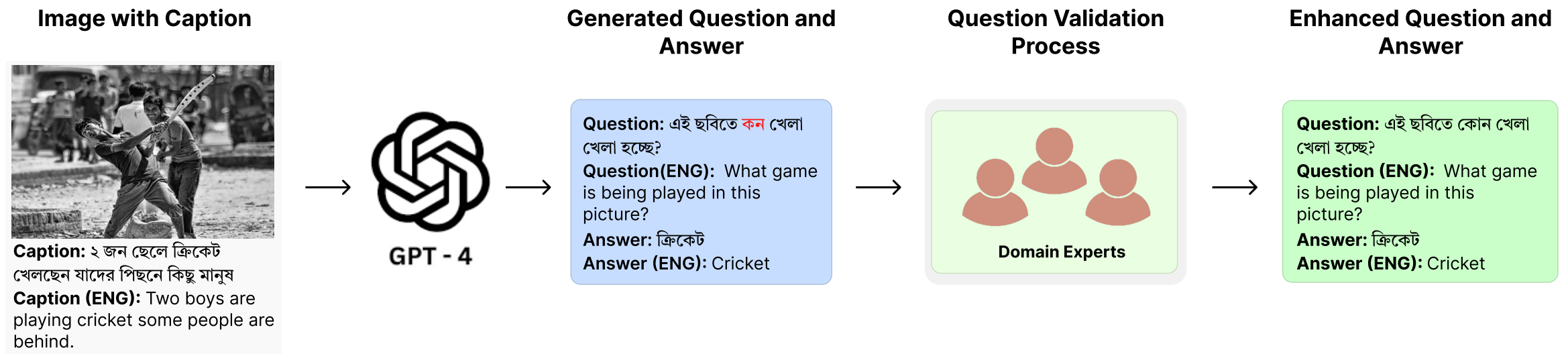

Figure 1: **Overview of the dataset generation pipeline.** The image-caption pairs are passed to GPT-4 Turbo using a curated prompt to generate QA pairs, then validated and corrected by the domain experts.

### 2.3 QA Pair Generation

We use OpenAI GPT-4 Turbo for the question-answer pair generation. The generated questions are comparatively more complex, diverse, and excel in other criteria, as seen in App. B. The prompts, reported in App. C.1, have been designed to constrain the answers to be within one to three words, ensuring that they succinctly capture what the question is asking without providing irrelevant information.

| **Source Distribution** | | |
|---|---|---|
| BanglaLekhaImageCaptions | | 8600 |
| Bornon | | 4292 |
| BNATURE | | 2400 |
| **Splits** | | |
| Train | | 12231 |
| Validation | | 1529 |
| Test | | 1532 |
| **General Statistics** | | |
| Samples | | 15292 |
| Captions | | 14927 |
| Images | | 15147 |
| Questions | | 13299 |
| Answers | | 5542 |
| WH-words | | 11 |
| Categories/Types | | 17 |
| **Q&A Statistics** | **Q** | **A** |
| Mean character length | 33.50 | 7.10 |
| Max character length | 105 | 45 |
| Min character length | 11 | 1 |
| Mean word count | 5.86 | 1.43 |
| Max word count | 17 | 4 |
| Min word count | 3 | 1 |

Table 2: Dataset statistics of ChitroJera.

| **QA Pair LLM Selection** | | |
|---|---|---|
| Samples used | | 1000 |
| Cohen's kappa coefficient | | 89.27% |
| **QA Pair Validation** | | |
| Samples used | | 2500 |
| | **Acc. (#)** | **Acc. (%)** |
| Annotator 1 | 2462 | 98.48 |
| Annotator 2 | 2476 | 99.04 |

Table 3: Data annotation statistics of ChitroJera. Accuracy refers to the correctness of the QA pairs.

### 2.4 Dataset Annotation

We verify the quality and consistency of the synthesized QA pairs by defining a few evaluation criteria, namely caption-question alignment, image-question alignment, question correctness, and answer correctness; discussed thoroughly in App. E. Our approach to data validation is methodological. We employ two experts in Bangla linguistics to evaluate the dataset on the aforementioned criteria over a set of 2500 randomly selected samples. Both experts are native Bangla speakers, accredited in linguistic proficiency, possess strong cultural awareness, and conform to the same principles in assessing grammatical correctness. We hired them on an hourly basis. The rationale for hiring domain experts instead of general annotators lies in their deeper understanding of the linguistic nuances of Bangla, enabling them to uphold the quality of the dataset. Since the overhead of cross-checking between two annotators is low, any inter-annotator disagreement is disputed directly via discussion. If no consensus is reached, we discard the ambiguous sample. However, such cases are quite rare, with only 5 samples being flagged as such. The figures for the annotation statistics are shown in Table 3.

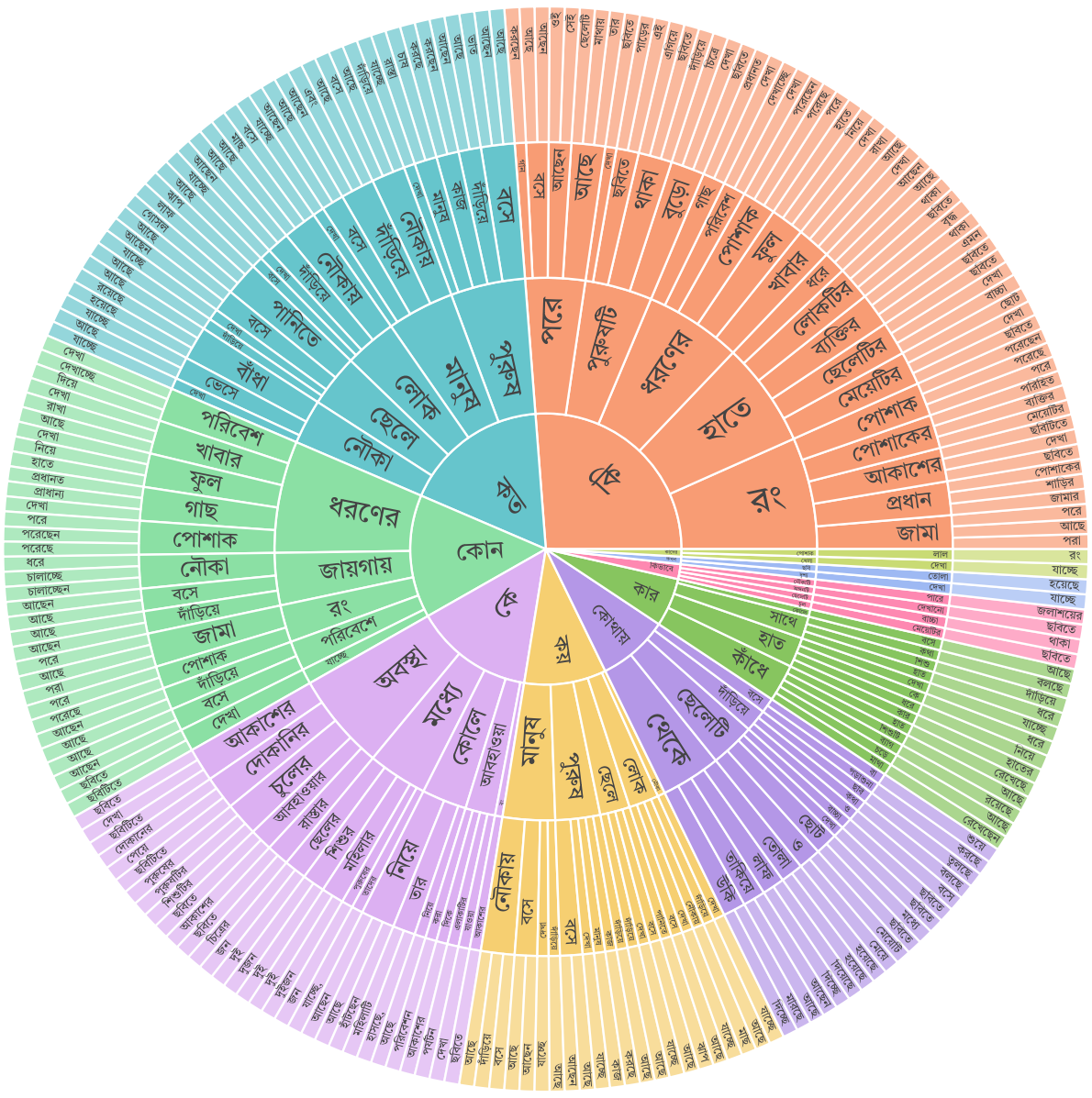

Figure 2: Question keyword distribution of ChitroJera.

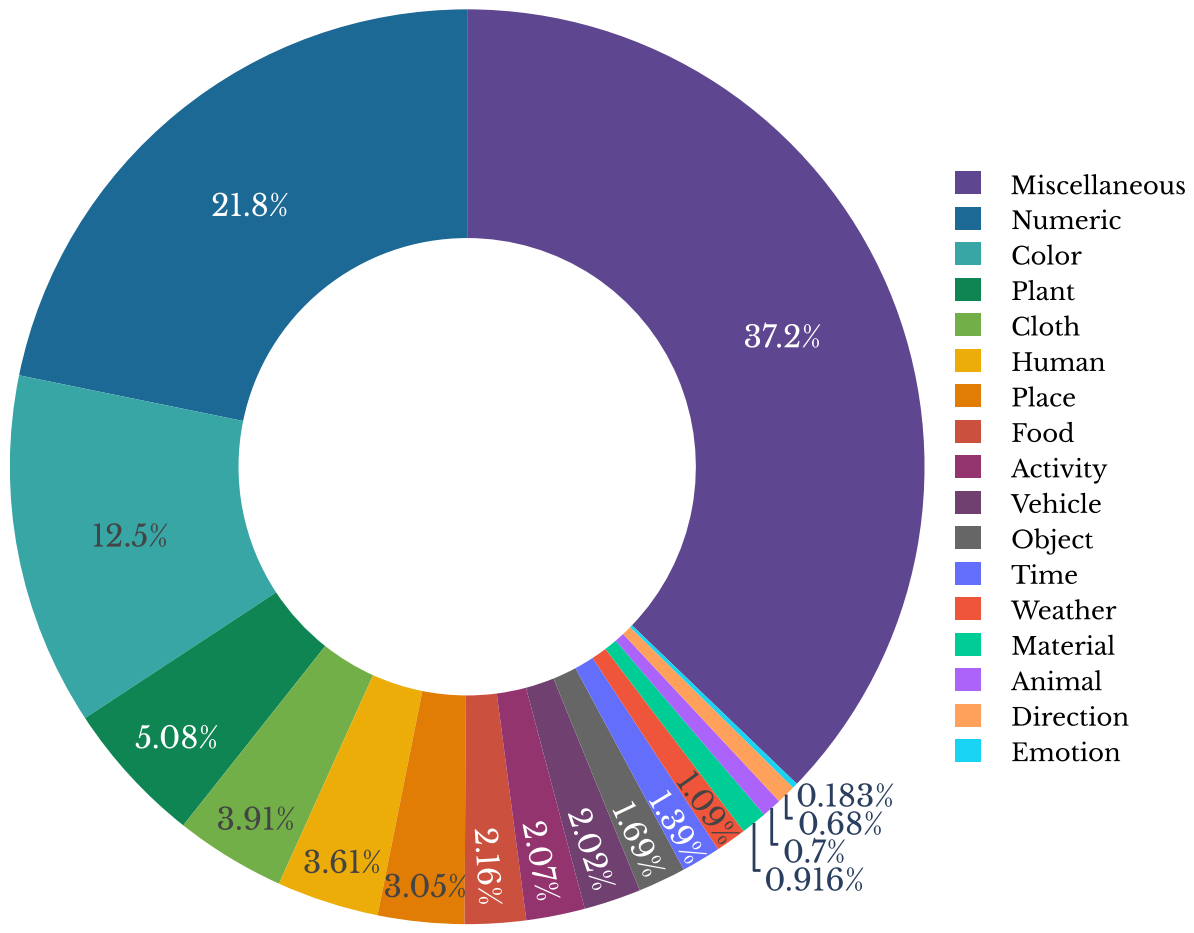

Figure 3: Answer category distribution of ChitroJera.

### 2.5 Dataset Statistics

We randomly split the dataset into training, validation, and test sets with a ratio of 80:10:10. To capture a more diverse representation of different contexts, we restrict the number of questions for each image to one or, at most, two. The number of unique questions is lesser than its total number, i.e. the same question has been asked on different images. This enables the models to generalize effectively, even when the visual context significantly differs. While the ratio of unique questions to unique answers is approximately 2.4, it does not necessarily indicate a sparsity in the training distribution. This is due to the inclusion of suffixes in the answers, i.e. the same root word can take multiple forms. The statistics of ChitroJera are shown in Tab. 2.

**Question Statistics.** The questions generated by GPT-4 Turbo range from a minimum of 3 words to a maximum of 17 words, with a mean of 6 words. To assess the diversity of the questions, we chose a set of 11 wh-question words in Bangla that comprehensively cover all the grammatically correct questions. These are — "কি" (tag questions), "কোন" (which), "কত" (how many/much), "কোথায়" (where), "কয়" (how many/much), "কে" (who), "কার" (singular whose), "কখন" (when), "কিভাবে" (how), "কাদের" (plural whose) and "কবে" (when). Figure 2 offers an overview of the distribution of the keywords respectively. We hypothesize that the low sample count of questions using "কার" (singular whose), "কখন" (when), "কাদের" (plural whose), and "কবে" (when) is due to the difficulty in assessing temporal and possessive qualities from still images and captions that are unassuming of its subjects.

**Answer Statistics.** The prompt used to generate the QA pairs (seen in App. C.1) limits the answers to short phrases. Hence, the answers are mostly single-word answers with some being single-character answers involving Bangla numerals. For gauging the distribution of the answers in terms of content, we employ a keyword-based method to classify them into the discrete and exclusive categories of "Numeric", "Food", "Place", "Weather", "Animal", "Color", "Plant", "Material", "Activity", Emotion", "Cloth", "Direction", "Human", "Vehicle", "Time" and "Object"; outlined in Figure 3.

## 3 Baselines

We establish different baselines on our dataset considering image modality, text modality, and both modalities along with LLMs and V-LLMs. We consider accuracy and BERTScore (Zhang et al., 2019) as our metrics. The experiment setup, implementation details, and model configurations used in creating baselines are reported in Appendix C.

### 3.1 Unimodal Fine-tuning

We analyze the impact of the text and image modalities of our dataset separately by fine-tuning a specific set of models. Following Marino et al. (2019); Schwenk et al. (2022), the problem is treated as a multi-class classification task, where each class represents a possible answer, with the loss being cross-entropy loss.

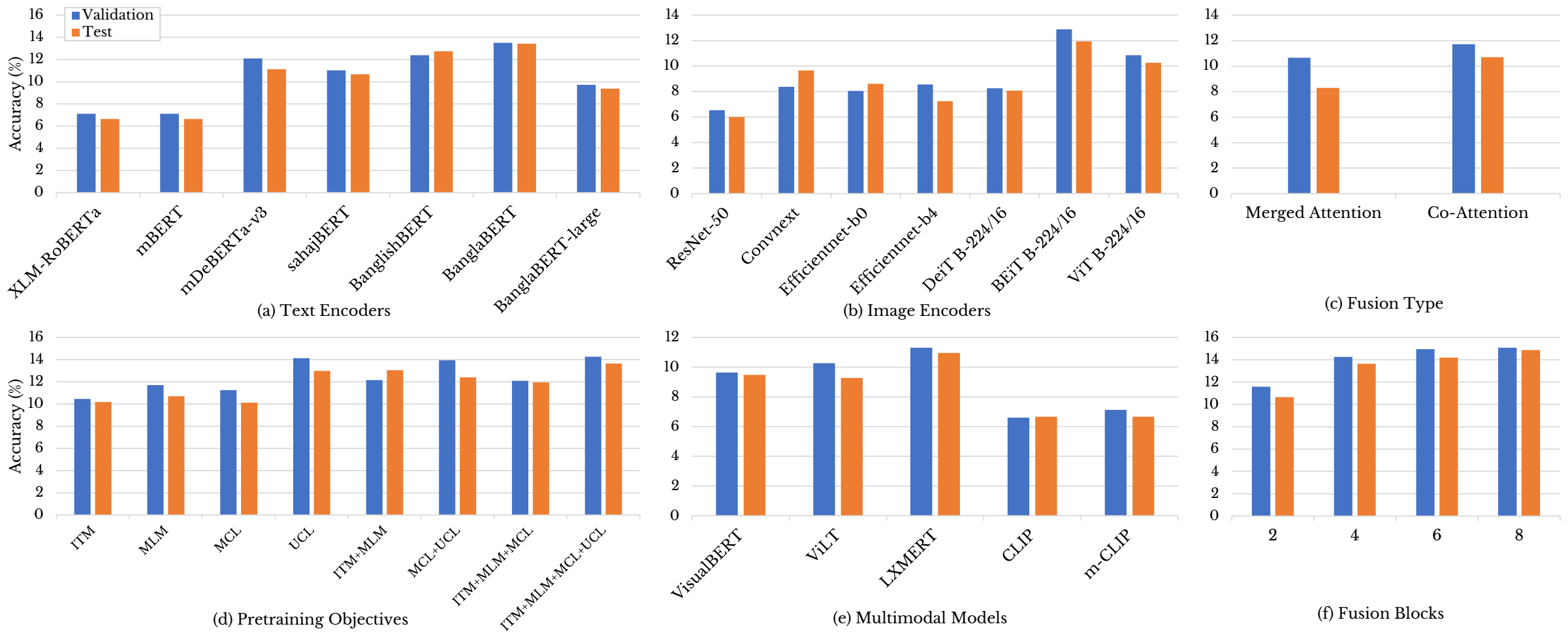

Figure 4: Validation and test accuracy across different text encoders, image encoders, multimodal models, fusion types, number of fusion blocks, and pretraining objectives.

In the text modality, we combine the question $q$ with its corresponding caption $c$ to define the combined input $x_t$ where $x_t$ = *Ques:* `q [SEP]` *Caption:* `c`. This combined input $x_t$ is then fed into a text encoder $f_T$ and then an MLP classifier head to get the predictions $\hat{y} = MLP(f_T(x_t))$. Only the `[CLS]` token representation is taken from the feature extractor. We consider multilingual models like XLM-Roberta (Conneau et al., 2019), mBERT (Libovický et al., 2019), and mDeBERTa-v3 (He et al., 2021), and Bangla language models such as sahajBERT[2], BanglishBERT and BanglaBERT (Bhattacharjee et al., 2021) as our dataset is in Bangla. Following Fig. 4a BanglaBERT exhibits the best performance.

For the visual modality, we pass the input image $x_{img}$ into an image encoder $f_I$ followed by MLP classifier head to get the predictions $\hat{y} = MLP(f_I(x_{img}))$. In the experiments, ResNet (He et al., 2016), Convnext (Liu et al., 2022), Efficientnet, (Tan and Le, 2019), DeiT (Touvron et al., 2021), BEiT, (Bao et al., 2021), and ViT (Dosovitskiy et al., 2020) are considered, where we observe BEiT and ViT outperforming the others (Fig. 4b).

### 3.2 Multimodal Fine-tuning

For addressing both the image and the text modality simultaneously, we also finetune with different VLMs, such as VisualBERT (Li et al., 2019), ViLT (Kim et al., 2021), LXMERT (Tan and Bansal, 2019), CLIP (Radford et al., 2021), and m-CLIP (Chen et al., 2023). In this scenario, the combined text input $x_t$ and the image input $x_{\text{img}}$ are processed by a VLM encoder $f_{VL}$. This yields separate text representation $\mathbf{h}_T$ and the image representation $\mathbf{h}_I$, or combined representation $\mathbf{h}$, expressed mathematically as follows:

$$\begin{aligned} \text{VisualBERT, ViLT:} \quad & \mathbf{h} = f_{VL}(x_t, x_{\text{img}}) \\ \text{LXMERT, CLIP, m-CLIP:} \quad & \mathbf{h}_T, \mathbf{h}_I = f_{VL}(x_t, x_{\text{img}}) \end{aligned}$$

Next, we extract the [CLS] token representations, denoted as $h_T^{[\text{CLS}]}$ and $h_I^{[\text{CLS}]}$. For the classification task, we concatenate these two representations $h_{\text{concat}} = [h_T^{[\text{CLS}]}; h_I^{[\text{CLS}]}]$. This concatenated representation is then passed through a multilayer perceptron (MLP) to obtain the logits for classification. For VisualBERT and ViLT we simply pass $h^{[\text{CLS}]}$ to the MLP for classification. The performance of these models has been shown in Figure 4e, where LXMERT outperforms others, but still falls behind unimodal models like BanglaBERT. We hypothesize that this is due to the lack of exposure to Bangla text in their training data, limiting their ability to align both modalities.

### 3.3 Pretrained Dual Encoders

From the discussion in Section 3.2, pre-trained multimodal models are typically trained on image-English text pairs, resulting in a limited semantic understanding of Bangla text. To address this gap, we adopt a dual-encoder approach to extract semantic features. However, these unimodal models initially lack alignment between modalities, which we address by training a multi-modal fusion network on a pretraining dataset with specific pretraining objectives.

[2] `https://github.com/tanmoyio/sahajbert`

Following the experimental results of Sec. 3.1, we utilize the best-performing models from each modality: BanglaBERT for text, and both BEiT and ViT separately for the image. For pretraining datasets, we consider the image-caption dataset to align the dual encoders for both modalities. As described in Section 2.1, we use BanglaLekhaImageCaptions, Bornon, and BNATURE to generate our VQA dataset. Therefore, we omit these datasets for our pretraining tasks and instead use the BanCap (Khan et al., 2022) dataset.

### 3.3.1 Pretraining Mechanism

During the pretraining stage, the image $x_{img}$ and its corresponding captions are fed $x_t$ into image encoder $f_I$ and text encoder $f_T$ separately, to extract image representations $\mathbf{h}_I$ and text representations $\mathbf{h}_T$ where $\mathbf{h}_I = f_I(x_{img}); \mathbf{h}_T = f_T(x_t)$. Here, $\mathbf{h}_I \in \mathbb{R}^{n_v \times d_v}$ and $\mathbf{h}_T \in \mathbb{R}^{n_t \times d_t}$, where $n_v$ and $n_t$ are the number of visual and textual tokens, respectively, and $d_v$ and $d_t$ are their dimensionalities. To align the image and text representations, a fusion module is employed. We explore two types of fusion modules: merged attention and co-attention like (Dou et al., 2022; Hendricks et al., 2021). The details about the experimental setup can be found in Appendix C.2.

**Merged Attention:** Merged attention integrates information from both modalities into a single attention mechanism. We concatenate both representations $\mathbf{h}_{VL} = [\mathbf{h}_I; \mathbf{h}_T] \in \mathbb{R}^{(n_v+n_t)\times d}$. $\mathbf{h}_{VL}$ is passed into $B$ different fusion blocks which is typically the transformer-encoder blocks (Vaswani et al., 2017).

**Co-Attention:** In the co-attention module, the features are processed through $D$ separate transformer blocks, utilizing techniques like cross-attention for cross-modal interaction. For each block $i$, the representations are calculated as follows:

$$
\begin{aligned}
Q_I^i, K_I^i, V_I^i &= W_{Q_I}^i \mathbf{h}_I^i, W_{K_I}^i \mathbf{h}_I^i, W_{V_I}^i \mathbf{h}_I^i \\
Q_T^i, K_T^i, V_T^i &= W_{Q_T}^i \mathbf{h}_T^i, W_{K_T}^i \mathbf{h}_T^i, W_{V_T}^i \mathbf{h}_T^i \\
\mathbf{h}_I^i &= \text{Self-Attn}(Q_I^i, K_I^i, V_I^i) \\
\mathbf{h}_T^i &= \text{Self-Attn}(Q_T^i, K_T^i, V_T^i) \\
\mathbf{h'}_I^i &= \text{Cross-Attn}(Q_I^i, K_T^i, V_T^i) \\
\mathbf{h'}_T^i &= \text{Cross-Attn}(Q_T^i, K_I^i, V_I^i) \\
\mathbf{h}_I^{i+1} &= \text{MLP}(\mathbf{h'}_I^i); \mathbf{h}_T^{i+1} = \text{MLP}(\mathbf{h'}_T^i)
\end{aligned}
$$

**Pretraining Objectives:** We utilize four different losses namely Masked Language Modeling (MLM), Image-Text Matching (ITM), Multimodal Contrastive Loss (MCL), and Unimodal Contrastive Loss (UCL), which are the most common and widely used pretraining objectives in the multimodal domain. From Figure 4d, combining all pretraining objectives as $\mathcal{L}_{PT} = \mathcal{L}_{ITM} + \mathcal{L}_{MLM} + \mathcal{L}_{MCL} + \mathcal{L}_{UCL}$. Detailed ablation study on pretraining objectives and their descriptions are presented in Appendix D.1.

| Dual Encoders | Performance Metrics | | | |
|---|---|---|---|---|
| | Validation | | Test | |
| | Acc. | BScore | Acc. | BScore |
| **w/o Pretraining** | | | | |
| BanglaBERT | | | | |
| + ViT-224 | 11.45 | 89.01 | 10.64 | 88.71 |
| + BEiT-224 | **12.68** | **89.87** | **11.11** | **89.26** |
| **with Pretraining** | | | | |
| BanglaBERT | | | | |
| + ViT-224 | 14.26 | **90.78** | **13.64** | **90.59** |
| + BEiT-224 | **14.45** | 89.50 | 13.12 | 89.56 |

Table 4: Evaluation of dual-encoder pre-training with $\mathcal{L}_{PT}$ training objectives using accuracy and BERTScore.

**Fusion Type:** Following Fig. 4c, the Co-Attention fusion outperforms the Merged Attention fusion. Hence, we choose Co-Attention as the type of fusion for our further experiments.

**Effect of Pre-training:** Table 4 presents the performance results of the selected dual-encoders with and without pretraining. The table indicates that pretraining enhances the dual encoder accuracy by approximately 2-3% on both the validation and test datasets. Given that our pretraining dataset contains only 44k samples, the improvement is modest.

### 3.3.2 Feature Aggregation based Fine-tuning

When employing a co-attention-based network for modality fusion, we get image-aware text representations $\mathbf{h}'_L$ and text-aware image representations $\mathbf{h}'_I$ after pre-training. While fine-tuning for VQA classification, we extract the [CLS] token representations, denoted as $h'^{[\text{CLS}]}_T$ and $h'^{[\text{CLS}]}_I$. To derive the ultimate representation for classification, we explore two aggregation techniques.

- **Concat-based:** The final representation $z$ is

| Models | Performance Metrics | | | | | |
|---|---|---|---|---|---|---|
| | Validation | | | Test | | |
| | Acc | BScore | LAVE | Acc | BScore | LAVE |
| **Dual Encoder Fusion Type** | | | | | | |
| BanglaBERT-ViT [Concat] | 14.26 | 90.78 | 15.83 | 13.64 | 90.59 | 16.08 |
| BanglaBERT-BEiT [Concat] | 14.45 | 89.50 | 20.60 | 13.12 | 89.56 | 20.10 |
| BanglaBERT-ViT [Sum] | 14.08 | 89.71 | 18.88 | 13.61 | 89.32 | 23.46 |
| BanglaBERT-BEiT [Sum] | 14.03 | 89.76 | 18.58 | 14.45 | 90.22 | 20.44 |
| **Open Source LLMs [Monolingual]** | | | | | | |
| BLIP-2 | 11.74 | 87.29 | 19.74 | 10.35 | 86.92 | 19.71 |
| InstructBLIP | 5.29 | 70.25 | 6.68 | 5.67 | 71.14 | 7.24 |
| LLaVa-1.5-7B | 7.93 | 74.86 | 8.80 | 6.73 | 73.75 | 7.93 |
| LLaVA-OneVision-7B | 7.89 | 73.64 | 8.79 | 6.68 | 71.64 | 7.88 |
| **Open Source LLMs [Multilingual]** | | | | | | |
| PaliGemma-3B | 8.44 | 79.26 | 9.37 | 8.98 | 80.72 | 10.59 |
| Pangea-7B | 11.26 | 86.28 | 17.94 | 10.28 | 86.26 | 18.88 |
| Qwen2.5-VL-7B | 12.31 | 88.04 | 18.36 | 12.04 | 87.93 | 18.45 |
| Phi-3.5-Vision | 10.67 | 83.57 | 19.01 | 10.31 | 83.26 | 19.97 |
| InternVL2-8B | 11.98 | 87.33 | 17.22 | 11.24 | 87.01 | 16.85 |
| **Closed Weights/Source LLMs** | | | | | | |
| Gemini 2.0 Flash | 23.07 | 90.15 | **62.31** | 26.58 | 89.15 | **66.08** |
| Claude 3.7 Sonnet | 21.55 | 88.72 | 52.76 | 28.09 | 89.48 | 63.82 |
| GPT-4o | 31.58 | 92.01 | 56.28 | 30.22 | 91.79 | 58.54 |
| GPT-4 Turbo | **33.35** | **92.28** | 61.30 | **32.83** | **92.18** | 57.79 |

Table 5: The performance of dual-encoder models using different modality aggregation techniques and large language models on the ChitroJera dataset.

calculated as follows:

$$z = \texttt{MLP}([h'^{[\text{CLS}]}_T ; h'^{[\text{CLS}]}_I])$$

- **Summed-based:** In this case, the final representation $z$ is calculated using:

$$z = \texttt{MLP}([h'^{[\text{CLS}]}_T + h'^{[\text{CLS}]}_I])$$

This aggregated representation $z$ is then fed into the classification head followed by a linear layer for prediction.

### 3.4 LLM Prompting

We also investigate the performance of LLMs & V-LLMs using our dataset through different prompting techniques. We experiment with two types of prompting: caption-based and non-caption-based. In the non-caption-based prompting approach, we input an image $x_{img}$ along with a question $x_t$. In contrast, the caption-based prompting method concatenates the corresponding caption for $x_{img}$ from our dataset alongside the question in $x_t$. In both types of prompting, we provide the necessary instructions to complete the task, as reported in App. F. We focus exclusively on zero-shot prompting for various LLMs and V-LLMs. Following the results of App. D.2, we selected the GPT models for further experiments.

## 4 Benchmarking and Analysis

Following Table 5, our evaluation largely focuses on the effectiveness of large language models, and to a smaller extent, the performance of our pretrained dual-encoder models at different configurations.

**Dual Encoder Models:** As shown in Table 5, the concat-based fusion technique outperforms the sum-based approach, with BEiT paired with BanglaBERT yielding better performance than the BanglaBERT-ViT combination. Concat-based fusion produces a 0.18% and 0.42% increase in accuracy compared to their sum-based counterparts for BanglaBERT paired with ViT and BEiT encoders, respectively. Using the BEiT encoder produced a 0.19% increase in accuracy compared to ViT for concat-based fusion, but saw a 0.5% decline for sum-based fusion. Further details on the hyperparameters have been provided in the supplementary material.

**Monolingual vs. Multilingual LLMs:** It can be seen from the benchmarks that models that have an explicit focus on multilingualness, which are models that have been trained on multilingual data, perform better on our dataset compared to monolingual models, largely focused on English. The viability of the tokenizers in a monolingual or multilingual setting also plays an important role (Ali et al., 2024). Most of the monolingual models have sub-10% accuracy and LAVE scores, while nearly all the multilingual models make it past that.

**Closed Source LVLMs:** Among the closed source LVLMs, GPT family outperforms Gemini and Claude with a margin of around 10% in terms of accuracy. Besides accuracy, the performances of Gemini 2.0 Flash, Claude 3.7 Sonnet, GPT-4o, and GPT-4 Turbo are comparable in terms of BERTScore and LAVE, with Gemini having an edge in LAVE. For a rigid metric like accuracy, GPT-4 has a bigger advantage due to the exactness between the generated answers during QA pair synthesis and blind answer generation. In open-ended answer generation, metrics like accuracy do not paint the full picture (Mañas et al., 2024).

**Dual Encoder vs. Open Source:** Our dual encoder models outperform all open-source LVLMs, including multilingual ones, across all evaluation metrics. The best-performing model, BanglaBERT-BEiT[Concat], achieves 14.45% accuracy and 20.60% LAVE on the validation set, and 13.10% accuracy and 20.10% LAVE on the test set. These results surpass those of the top-performing zero-shot open-source LVLM, Qwen2.5 VL, which achieves 12.31% accuracy and 17.94% LAVE on the validation set, and 12.04% accuracy and 18.45% LAVE on the test set. While this comparison may not be entirely fair, it suggests that open-source LVLMs lack sufficient Bangla regional pre-training data, indicating the need for fine-tuning to improve performance.

**Open Source vs. Closed Source LVLMs:** Based on the benchmarks, the proprietary LLMs surpass the results of open-source models in every evaluation metric. Among the open-source models, Qwen2.5-VL-7B attains the best performance in accuracy and BERTScore. Phi-3.5-Vision, on the other hand, offers the best LAVE score, suggesting that while it does not get the answers as exact as Qwen2.5, it has a more holistic idea of what the correct answer might be, having considered semantic equivalence, synonyms, and variations in verbosity. Proprietary models, such as Gemini 2.0 Flash, Claude 3.7 Sonnet, GPT-4o, and GPT-4 Turbo, are trained on multilingual data. The scale of their training is reflected in the performance improvement over the open-source models.

## 5 Ablations and Hyperparameter Tuning

In the dual encoders, we use $M_{co_attn} = 4$ and $M_{merged_attn} = 8$ co-attention and merged attention modality fusion blocks respectively. Ablation on the number of modality fusion blocks $M_{co_attn}$ is illustrated in Fig. 4.

| Batch Size | Validation | | Test | |
|---|---|---|---|---|
| | **Acc.** | **BScore** | **Acc.** | **BScore** |
| 8 | 12.94 | 90.41 | 11.28 | 90.14 |
| 16 | 14.00 | 90.46 | 12.47 | 90.23 |
| 32 | 12.10 | 90.30 | 10.38 | 90.08 |
| **64** | **14.26** | **90.78** | **13.64** | **90.59** |

Table 6: Effect of batch size on the `concat-based` ViT-224 and BanglaBERT dual-encoder model, evaluated using accuracy and BERTScore.

As per Tab. 6, the accuracy and BERTScore show somewhat inconsistent improvement with the increase in batch size, with the best performance observed at size 64. Further details on the implementation and hyperparameters have been provided in App. C.2.

## 6 Error Analysis

From Appendix Fig. 11, it can be reasonably inferred that implicit answers can affect the V-LLM performance. In the first image, GPT-4o struggles with the ability to differentiate between two similar-looking objects, failing to tell a “টায়ার” (tire) apart from a “চাকা” (wheel) as both objects are circular. The second example showcases an example of Western bias where the model is unable to recognize the action of drying paddy through manual labor, a common practice in rural Bengal. Finally, using accuracy, as a metric, treats answers too rigidly, especially where contextual relevance is not assessed. In the third example, the captionless answer "সবজি বাজার" (vegetable market) is technically acceptable but is considered an error as the ground truth is "কাচা বাজার" (fresh market). For all three cases, adding the caption within the context makes the answers accessible in the textual domain and results in improved performance.

Categorically, Appendix Fig. 10a and 10b outline that GPT-4 Turbo exhibits inconsistent performance across all sets on the "Emotion" category, likely due to the low sample count in this category. Following appendix Fig. 10c and 10d, with the addition of captions to the context, the accuracy among different categories is largely maintained, with additional gains for the "Activity", "Animal", "Cloth", "Food" and "Weather" classes. While "Emotion" retains a similarly poor trend as before, the "Material" class benefits immensely from the additional information.

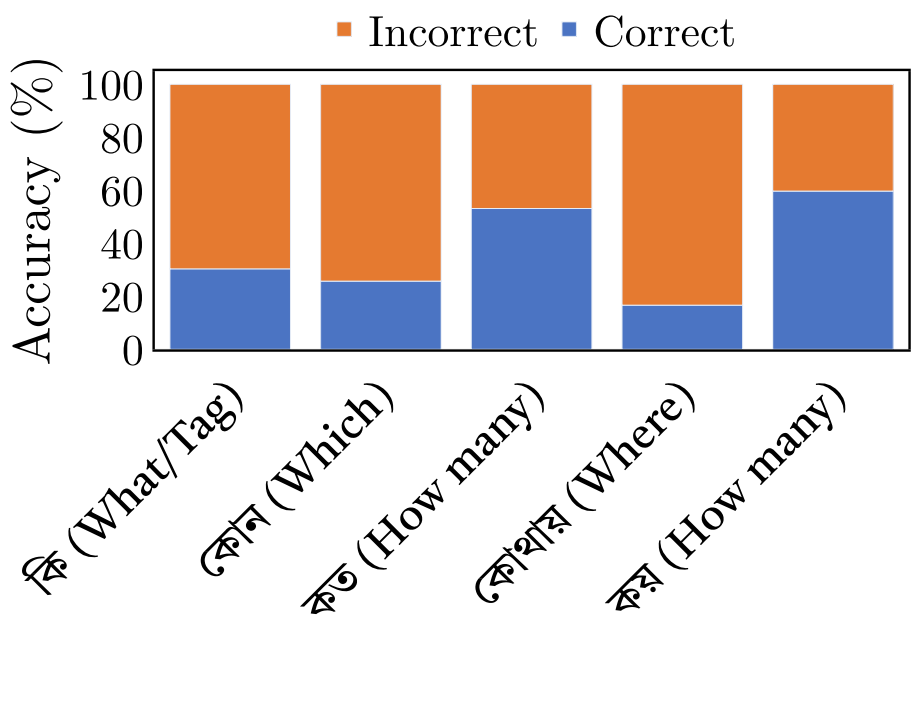

(a) Valid

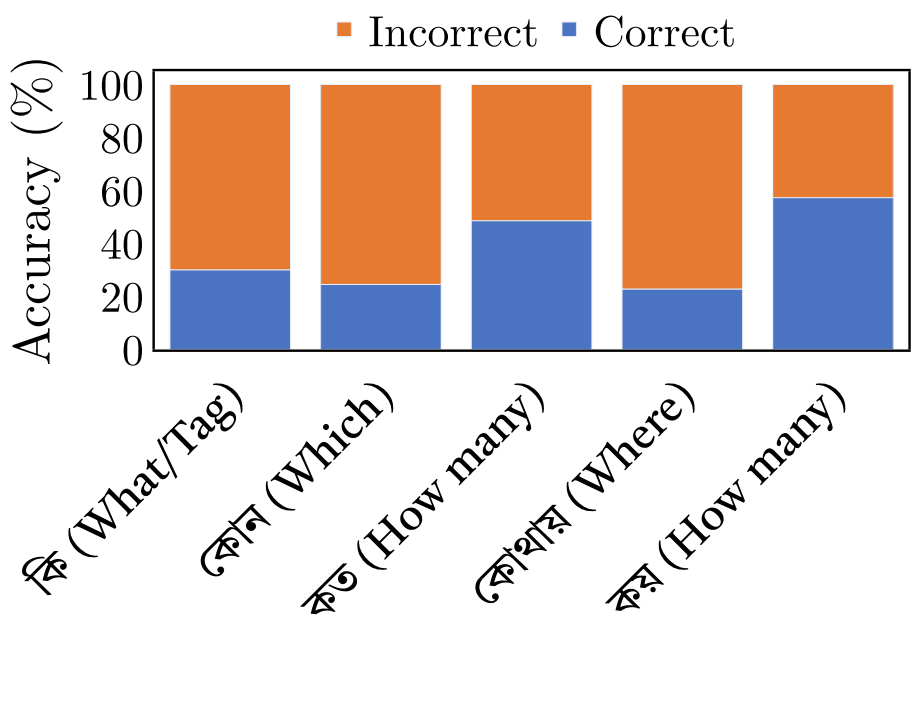

(b) Test

Figure 5: WH-word-based accuracy by GPT-4 Turbo on valid and test sets

For WH-words, GPT-4 Turbo attains 50% or more accuracy on counting questions on both test and validation splits, as seen in Fig. 5. For WH-words such as "কি" (what/tag question) and "কোন" (which), the accuracy is close to the model average. Performance is comparatively poor for "কোথায়" (where), as spatial reasoning from a still image can be challenging. These findings resemble the explicit/implicit subject trend from the categorical breakdown. Note that Fig. 5 only lists the top 5 WH-words (in terms of sample count) for brevity.

## 7 Related Work

**VQA Datasets and Models.** Visual Question Answering datasets have significantly evolved over the years (Ishmam et al., 2024), with early datasets like DAQUAR (Malinowski and Fritz, 2014) and COCO-QA (Ren et al., 2015) providing initial benchmarks at restricted settings. The VQAv1 dataset (Antol et al., 2015) introduced free-form questions on real and synthetic images, which became a foundational benchmark despite the criticisms of linguistic bias, later addressed by VQAv2 (Goyal et al., 2017). Visual Genome (Krishna et al., 2017) and Visual 7W (Zhu et al., 2016) further enriched the VQA landscape by incorporating a vast number of QA pairs and scene graphs. CLEVR (Johnson et al., 2017), GQA (Hudson and Manning, 2019), and the recent OK-VQA (Marino et al., 2019) datasets, have pushed the boundaries by incorporating reasoning, external knowledge, and challenging biases in traditional datasets.

VQA methods have transitioned from the traditional joint encoding schemes (Antol et al., 2015; Fukui et al., 2016) to Vision Language Pre-training (VLP) (Li et al., 2019; Zhou et al., 2020; Kim et al., 2021; Bao et al., 2022; Li et al., 2022). The foundation of VLP lies in the transformer architecture (Vaswani et al., 2017), which can capitalize pre-training on large visual and textual corpora that can be fine-tuned for specific downstream tasks like VQA.

**Multilingual and Bangla VQA.** The majority of the aforementioned VQA datasets are focused on the English language while multilingual VQA (Changpinyo et al., 2022) extends the VQA capabilities to other languages. Early efforts in creating multilingual datasets, such as Multi30K (Elliott et al., 2016) and xGQA (Pfeiffer et al., 2021), have involved translating existing English datasets, gen-erating synthetic multilingual data, and employing language-agnostic embedding to ensure comprehension of multiple languages. The associated models frequently relied on fine-tuning multilingual transformers like mBERT (Devlin et al., 2019) and XLM-R (Conneau and Lample, 2019) as modules for multilingual VQA systems.

Bengali-VQA-v1 and Bengali CLEVR (Islam et al., 2022) are derived from English VQA v1 (Antol et al., 2015) and CLEVR (Johnson et al., 2017) respectively, relying on machine translation. CVQA (Romero et al., 2024) takes a more refined approach by offering a VQA dataset of 26 languages built from the ground up with relevant cultural connections, with Bangla being one of them.

**LLMs in VQA Tasks.** Large language models have been employed in both traditional (Chappuis et al., 2022; Shao et al., 2023) and zero-shot settings (Guo et al., 2023; Liu et al., 2024a) of VQA. Multimodal Large Language Models (MLLMs) or Large Vision Language Models (LVLMs) (Zhang et al., 2024) like GPT-4 (Achiam et al., 2023) and LLaVA (Li et al., 2024) have shown great VQA performance in the zero-shot and few-shot settings. Our work evaluates both traditional and large VLMs on our dataset.

## 8 Discussion and Conclusion

We introduce ChitroJera, a VQA dataset deeply rooted in the geography, culture, and norms of the Bangla-speaking region. To our knowledge, ChitroJera is the first VQA benchmarking with images relevant to the Bengal region, filling a crucial gap in the Bangla vision-language landscape. Our novel dual-encoder models outperform existing unimodal and multimodal models of comparable scale, while V-LLMs, such as the GPT-4, attain superior performance primarily due to their scale and pre-training data. However, the performance gain due to the inclusion of captions outlines textual bias and the potential use of bias mitigation techniques in low-resource VQA. We anticipate that our work will foster improvements in future models, enabling better modality alignment with low-resource contexts and thereby, overall improved performance.

## Limitations

Our work primarily explored WH-word questions of different types, with some exhibiting traits that are harder to capture from still images. Additionally, large language models might have implicit biases that prefer certain WH-words more than others, which can be somewhat mitigated via robust prompt engineering techniques. Since WH-words do not account for all types of questions either, the same question can be asked differently for better generalization.

## Ethical Statement

Human annotators have been made aware of the full extent of the research and its potential outcomes. PII from image metadata and the content has been removed.

**LLM Use:** We acknowledge the implicit bias of the Large Language Models (LLMs) that have been used to generate the QA pairs of our dataset. LLMs have only been used to generate the synthesized dataset, correct grammatical errors, and polish written texts. The use of AI strictly adheres to the ACL AI Assistance policy[3].

**$CO_2$ Emission:** With a carbon efficiency of 0.432 kg$CO_2$eq/kWh (OECD average), a total of 150 hours of computation was performed using Tesla P100 hardware (TDP of 250W) for the unimodal, multimodal, and dual encoder models. Total emissions of those experiments are estimated to be 16.2 kg$CO_2$eq.

[3]https://2023.aclweb.org/blog/ACL-2023-policy/

**Algorithm 1** Select Captions and Form Textual Ground Truth Context

**Require:** Set of captions $C = \{c_1, c_2, \ldots, c_n\}$ for an image
**Ensure:** Textual ground truth context $T$
$longest \leftarrow argmax(len(c_i))$
$shortest \leftarrow argmin(len(c_i))$
$C' \leftarrow C \setminus \{longest, shortest\}$
$best_bs \leftarrow -\infty$
$best_caption \leftarrow$ None
**for all** $c \in C'$ **do**
    $bsl \leftarrow BERTScore(c, longest)$
    $bss \leftarrow BERTScore(c, shortest)$
    $bs \leftarrow bsl + bss$
    **if** $bs > best_bs$ **then**
        $best_bs \leftarrow bs$
        $best_caption \leftarrow c$
    **end if**
**end for**
$T \leftarrow longest +$ " " $+ shortest$
**if** $best_caption$ is not None **then**
    $T \leftarrow T +$ " " $+ best_caption$
**end if**
**return** $T$

## A Caption Filtering

Filtering has been used to form the context for the experiments with the captions included. If there are 3 or fewer captions, all of them are added to the context. However, for source images that have more than 3 captions available, filtering is required to remove redundant and repeated information. For this, we pick the longest and the shortest caption as these two are more likely to offer differing perspectives of the same image: the longer one captures more detail and the shorter one paints the big picture. Lastly, we also pick a third caption, the one with the closest BERTScore to the previous two captions. This captures slight differences in semantics and suffixation, if any. The algorithm that we use for caption filtering is shown in Algorithm 1.

## B LLM Selection

We select the LLM for QA pair generation by comparing the ones generated by GPT-4 Turbo against the ones generated by Google Gemini 1.5 Pro. This is performed over a sample of 1000 image-caption pairs, where two domain experts are asked to either pick GPT-4 Turbo or Gemini 1.5 Pro for each sample. Since there are exactly two annotators and all the criteria to be evaluated are categorical (binary), we find Cohen's kappa coefficient to be a suitable basis for quantifying the agreement between the two annotators, obtaining a score of 89.27%. GPT-4 Turbo is picked in most of the instances, making it favorable over Gemini 1.5 Pro for the question synthesis, as evident from the confusion matrix in Table 7.

| | **Ann. 2 (GPT-4)** | **Ann. 2 (Gemini)** |
|---|---|---|
| **Ann. 1 (GPT-4)** | 979 | 3 |
| **Ann. 1 (Gemini)** | 1 | 17 |

Table 7: Confusion matrix of annotator LLM preferences for QA pair generation (GPT-4 Turbo vs. Gemini 1.5 Pro).

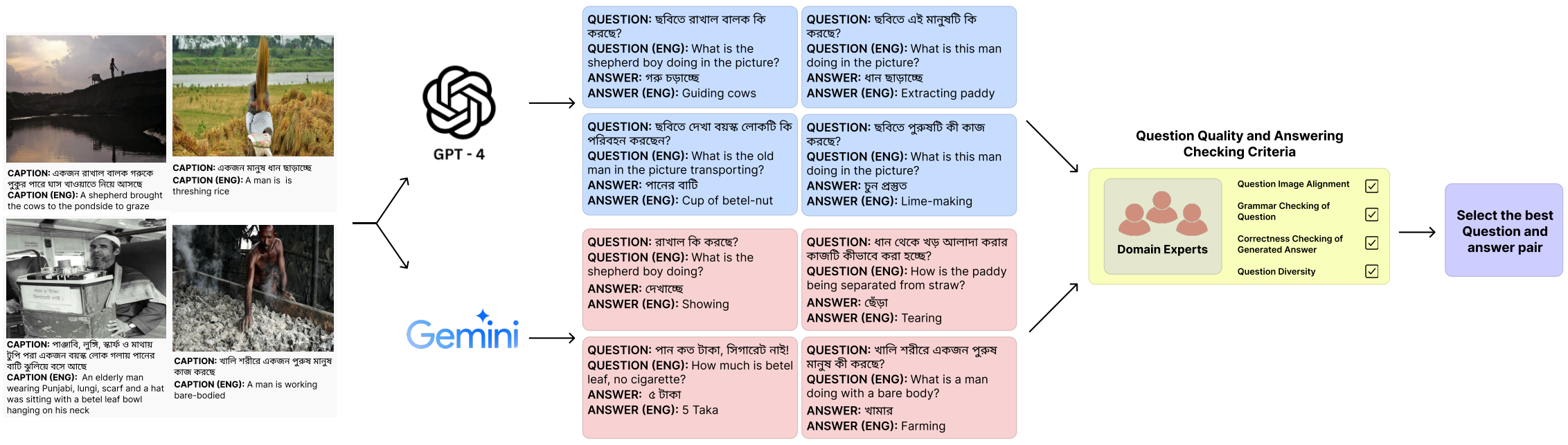

Figure 6: A flow of GPT-4 vs Gemini question answer generation and expert verification process.

# C Experiment Setup

## C.1 Prompt Design for Generating Question and Answers using LLMs

The prompt we used for generating questions and answers using LLMs by inputting captions and images is given below. We also set some guidelines so that the response from the LLMs aligns with our objectives.

**Prompt to Generate the Question-Answer Pairs for the dataset**

You are an expert in generating Bangla visual question answers. For a given image and the captions to the image, your task is to generate the question and the answer. You should always abide by the guidelines that are mentioned below:
GUIDELINE 1: The questions should be always image-aligned, caption-aligned, and informative
GUIDELINE 2: Try to generate the answer in one or two words. The answer must never contain more than three words
GUIDELINE 3: Generate the question-answer pair in the Bangla language
Here is the caption:
<CAPTION>
Based on the captions above and the image, generate one question-answer pair in Bangla. Generate the question-answer pair in the following format:

Q# <GENERATED QUESTION>, A# <GENERATED ANSWER>

## C.2 Implementation Details

All of the pretraining experiments were conducted using Python (version 3.12) and PyTorch on an NVIDIA P100 GPU. All pretraining models used in the paper were sourced from the Huggingface Transformers library (version 4.40.1), which we also utilized for fine-tuning the models. We trained the models for 15 epochs using the AdamW optimizer. The learning rate is $2 \times 10^{-5}$, the epsilon value of the optimizer is $10^{-6}$, and the beta coefficients for computing the running average of the gradient and its square are 0.9 and 0.999, respectively.

For all operations, the seed value used to ensure random state consistency was 42. The dimensions of the text and vision projection layers are 768. We set the dropout rate to 0.1, effectively zeroing 10% of the input values. The fine-tuning experiments were done using the same versions of Python and PyTorch, with two NVIDIA Tesla T4 GPUs. The dimension for the hidden layer of the classification head is 256 across all experiments. All other hyperparameters remained unchanged.

## C.3 Accuracy-BERTScore Mismatch.

Suffixes play a major role in Bangla, with the most common ones being “টা” or “টি” for inanimate objects and “জন” for for enumerating people. As observed across all instances, the BERTScore is noticeably higher compared to the accuracy. This is because accuracy only accounts for exact matches which includes suffixes, whereas BERTScore also accommodates semantically close but not exact matches.

In Table 4, there are instances where the accuracy of a model is higher compared to the rest but the BERTScore is not, and vice-versa. This is because a change in accuracy does not necessarily guarantee an equivalent change in the BERTScore. For reference, the BERTScore between “তিন” and “তিনজন” is 0.95, whereas it is 0.96 between “তিন” and “তিনটি”. If the ground truth is “তিন”, the sample is misclassified in both the cases, and their accuracies will be the same (0), but the BERTScores will differ (0.96 & 0.95).

## D Pretraining Objectives and Ablation Study

### D.1 Pretraining Objectives

- **MLM:** Masked Language Modeling (MLM) (Devlin et al., 2019; Zhuang et al., 2021; Dou et al., 2022) involves randomly masking some input tokens in a sentence from an image-caption pair. The model is then trained to reconstruct the original tokens using the masked tokens and the corresponding visual input.

- **ITM:** Image-Text Matching (ITM) (Li et al., 2022) requires the model to identify which image-caption pairs in a batch are correctly matched and which are not.

- **MCL:** Multimodal Contrastive Loss (MCL) (He et al., 2020; Li et al., 2023b)focuses on learning the alignment between the two modalities by maximizing the similarity between images and their corresponding text captions while distinguishing them from negative examples.

- **UCL:** Unimodal Contrastive Loss (UCL) (Li et al., 2023b) aims to differentiate examples within a single modality, such as images or text, in latent space to ensure similar examples are close together.

### D.2 Zero-shot baselines of LLMs

| Prompting | Performance Metrics | | | |
|---|---|---|---|---|
| | Validation | | Test | |
| | Acc. | BScore | Acc. | BScore |
| BLIP-2 | 11.20 | 87.53 | 11.78 | 87.64 |
| LLaVa | 18.90 | 90.34 | 17.18 | 90.33 |
| Gemini | 19.80 | 90.28 | 20.90 | 90.06 |
| **GPT-4** | **34.50** | **92.81** | **32.36** | **92.25** |
| GPT-4o | 32.30 | 92.32 | 30.26 | 91.85 |

Table 8: Performance of different prompt-based VLMs and Multimodal LLMs on 1000 samples of the validation and test set using accuracy and BERT score.

We assess the performance of several large vision-language models, including BLIP-2 (Li et al., 2023a), LLaVa (Liu et al., 2024b), and large language models like Gemini (Team et al., 2023), GPT-4, and GPT-4o (Achiam et al., 2023), on our dataset. This evaluation involves employing prompt-based techniques, where these models are prompted to predict answers based on the given image and question. For this experiment, we used 1,000 samples from both the validation and test datasets. The prompt used is detailed in the Appendix. The results, shown in Table 8, indicate that GPT models consistently outperform other large VLMs and LLMs. Therefore, we selected GPT models for benchmarking on our dataset.

## E Dataset Validation Criteria

- **Caption-question alignment:** This validates whether the question is aligned with the caption or not, ensuring that GPT-4 did not augment extraneous and irrelevant information to the generated question by hallucinating. The annotators remove any instance of extraneous information, if present.

- **Image-question alignment:** This is to validate if the question corresponds to the image or not. This prevents the samples from having errors due to invalid interpretation of the visual context and GPT-4's inability to detect the subjects that are not apparent in some instances. The annotators handle any references to erroneous information from the question.

- **Question correctness:** This metric gauges whether the question is grammatically correct or not. Despite having resilience in terms of grammar in general, GPT-4 may have trouble generating proper structure for low-resource languages such as Bangla. Hence, the grammatical correctness is verified and enforced.

- **Answer correctness:** This is to check if the answer generated by GPT actually resolves the question or not. If the response doesn't fully or partially address the question, it is revised.

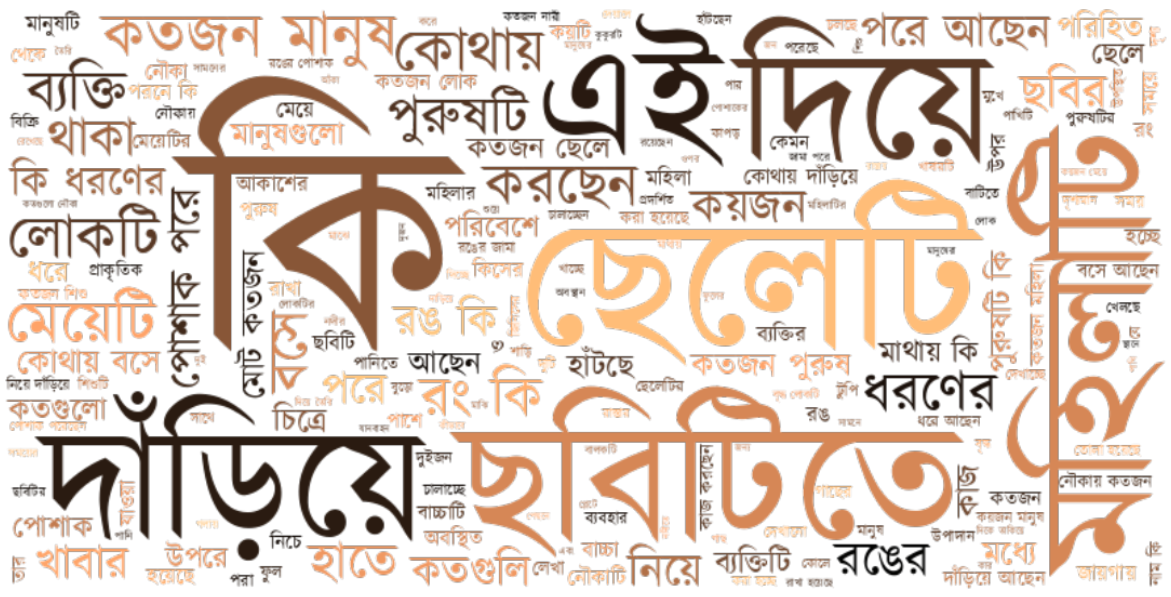

(a) Valid set (Bangla)

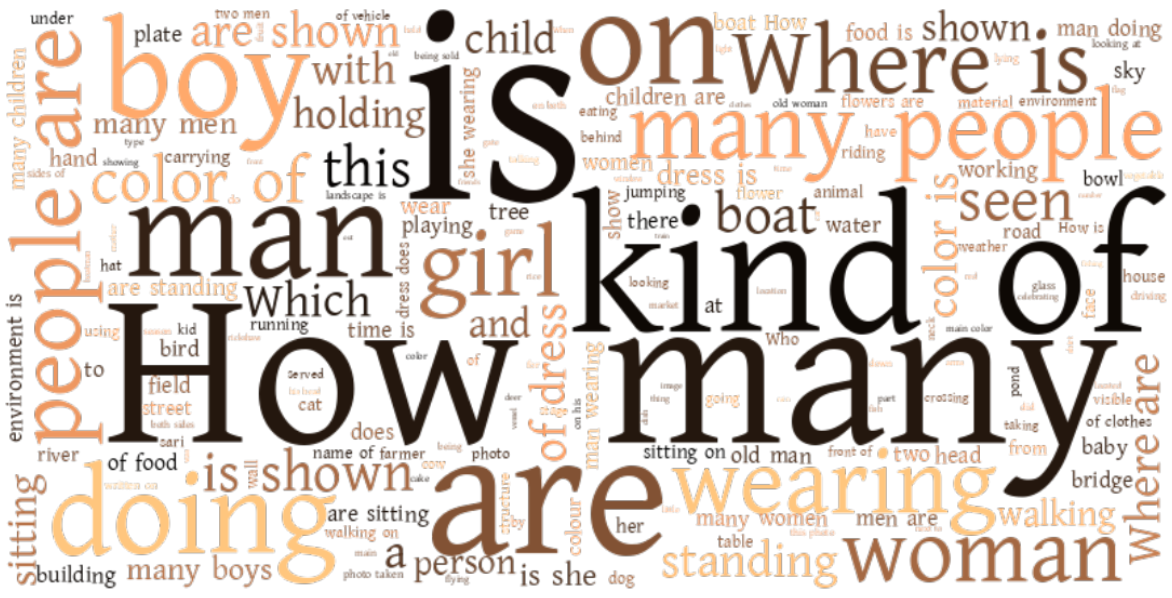

(b) Valid set (English)

Figure 7: Bangla and English word clouds of the questions from the validation set.

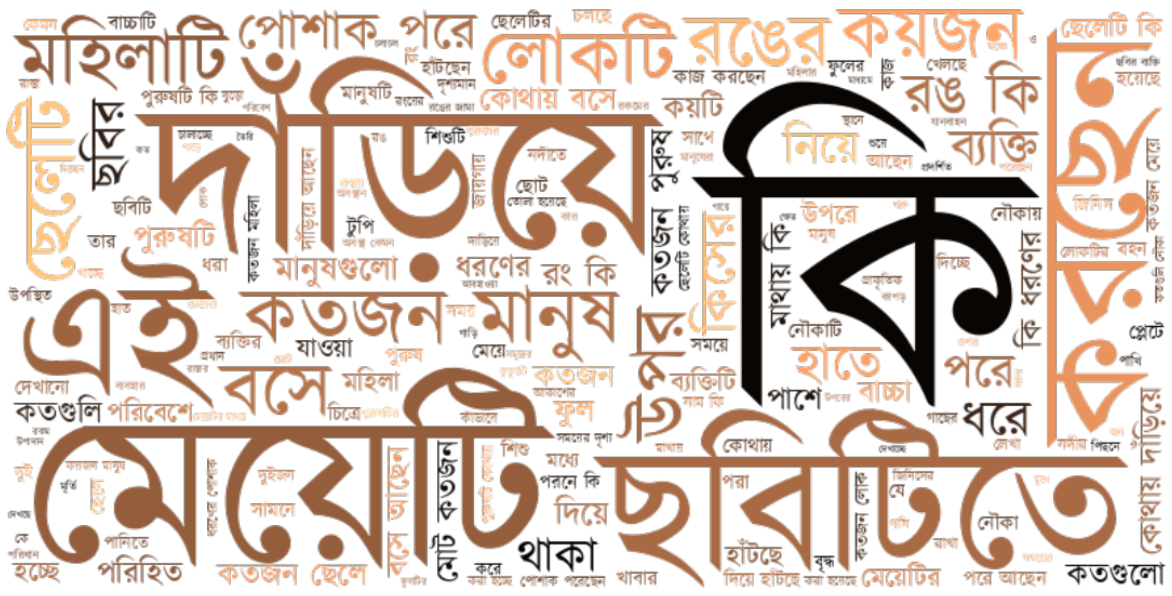

(a) Test set (Bangla)

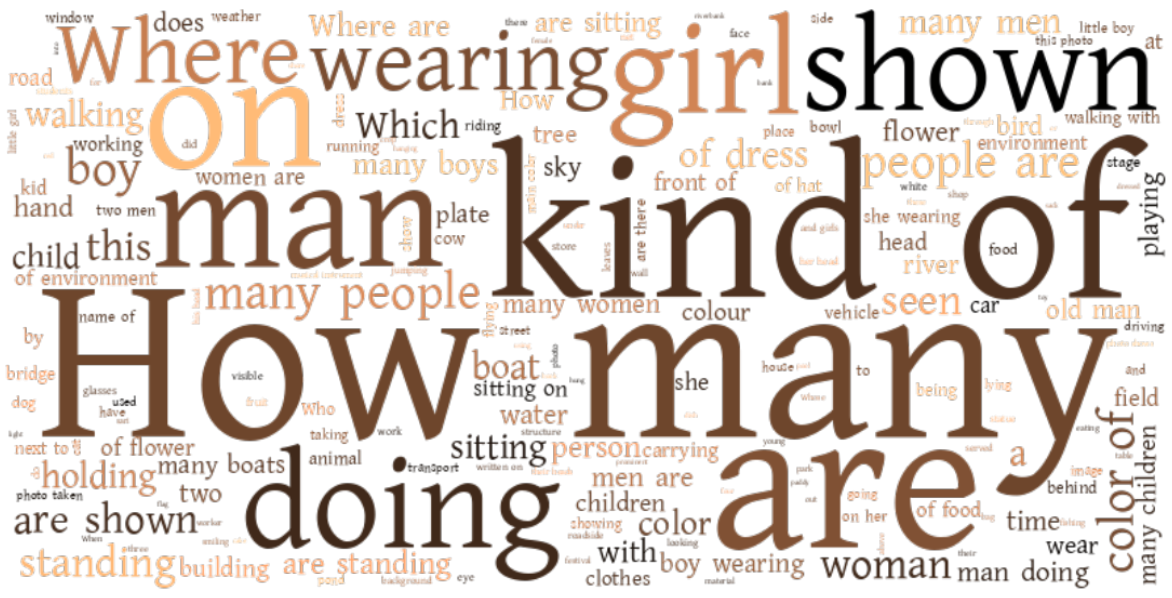

(b) Test set (English)

Figure 8: Word cloud of the Questions of test set in Bangla and English

## F Prompts used for LLMs for Prediction in our Dataset

**Prompt to Generate Answers using GPT-3.5 (Text only model)**

You are an expert Bangla question answering assistant. Given a caption, when asked a question with the context of the caption, your task is to generate the answer. You should always abide by the guidelines that are mentioned below:
1: Try to generate an answer of one or two words. The answer must never contain more than three words.
2: Always answer the question in Bangla language.
<CAPTION>,
<QUESTION>
When generating the Bangla answer of the question mentioned, generate in the following format:
A# <GENERATED ANSWER>

**Prompt to Generate Answers with caption**

You are an expert Bangla visual question answering assistant. Given an image with its caption, your task is to generate the answer when asked a question with the context of the image. You should always abide by the guidelines that are mentioned below:
1: Try to generate an answer of one or two words. The answer must never contain more than three words.
2: Always answer the question in Bangla language.
<CAPTION>,
<QUESTION>
When generating the Bangla answer of the question mentioned, generate in the following format:
A# <GENERATED ANSWER>

**Prompt to Generate Answers without caption**

You are an expert Bangla visual question answering assistant. Given an image, your task is to generate the answer when asked a question. You should always abide by the guidelines mentioned below:
1: Try to generate an answer of one or two words. The answer must never contain more than three words.
2: Always answer the question in Bangla language.
<QUESTION>
When generating the Bangla answer of the question mentioned, generate in the following format:
A# <GENERATED ANSWER>

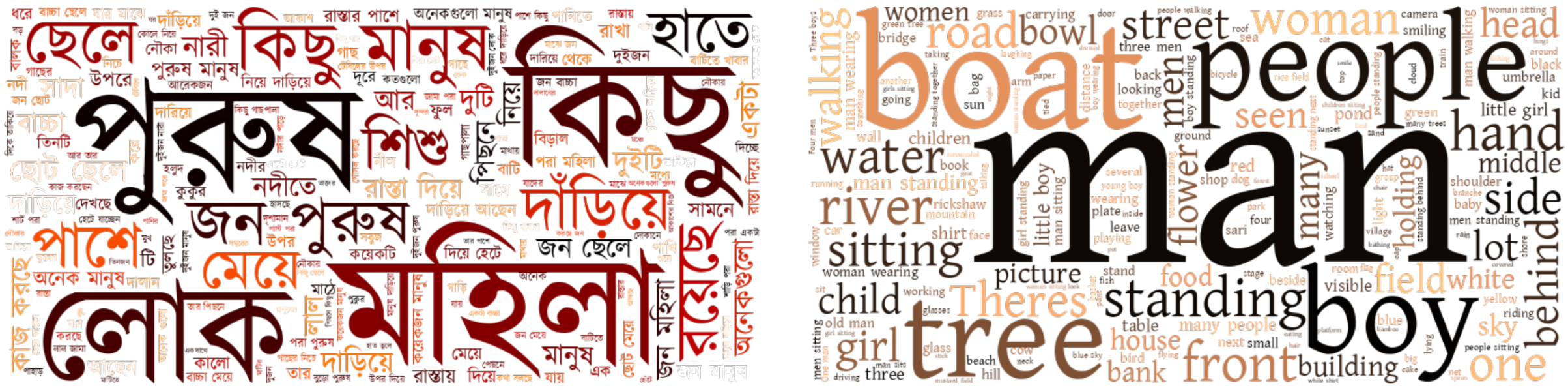

(a) Full dataset (Bangla)

(b) Full dataset (English)

Figure 9: Word cloud of the Captions of full dataset in Bangla and English

## G Word clouds

We also present word clouds for the question and answer texts across the train, validation, test, and overall datasets. These visualizations provide a clear overview of the most frequent terms, helping to identify common themes and vocabulary patterns within each subset of the data. From the Figures 7a and 8b, it can be observed that the validation and test splits of the datasets have similar distributions, with words like "কি" (tag question), "দাঁড়িয়ে" (standing), and "ছবিটিতে" (in the picture) having a large presence on both. Their English counterparts translate a bit differently due to the lack of certain kinds of suffixes, with phrases such as "How many", "kind of" and "man" being prevalent. These words and phrases have multiple representations in Bangla, causing them to split across multiple smaller entities.

As opposed to the word clouds of the questions, the word clouds of the captions in Figure 9 have a noticeably different distribution, as they lack inquisitive words. This also suggests why the performance improves when the caption is added to the context, as new information not present in the questions is being added.

## H Additional Error Analysis

The experiment results suggest that even with the information used to create the questions being accessible to the models verbatim for zero-shot prompts, they struggle to get the answers correct across all the instances. The degree of error varies, but with none of the options achieving 70% or more accuracy, there is much room to analyze exactly where these models are performing poorly.

### H.1 Category-wise Accuracy Score

From Figure 10, it can be inferred the some of the categories, such as "Color", "Human", "Number", "Object" and "Vehicle" perform consistently well across both the validation and test splits, with and without captions. All of these categories have one thing in common, that they can be inferred with ease from the images alone as their visual presence is explicit. On the other hand, the more abstract or implicit the category is, the harder it gets for the LLM to reason from the visual information alone. This is reflected in categories such as "Emotion", "Direction" and "Time".

### H.2 Caption Error Analysis

As reflected in Figure 11, the inclusion of captions in the context improve performance across all the models to some extent. GPT 3.5 seems to have a poorer understanding of Bangla expressions, as it answers the third question about the type of the market with an incomplete answer, where it fixates too much on the modifier instead of using the whole phrase coherently. GPT-4 and GPT-4o, on the other hand, obtain modest accuracy despite the metric requiring exact matches. This insinuates that the models answering incorrectly only using the image can either be attributed to the lack of knowledge regarding exact way the answer was used with the exact same suffixes, or being wrong entirely. The latter case can take place when the image does not provide enough fidelity, or when the cultural context is not immediately available to the model.

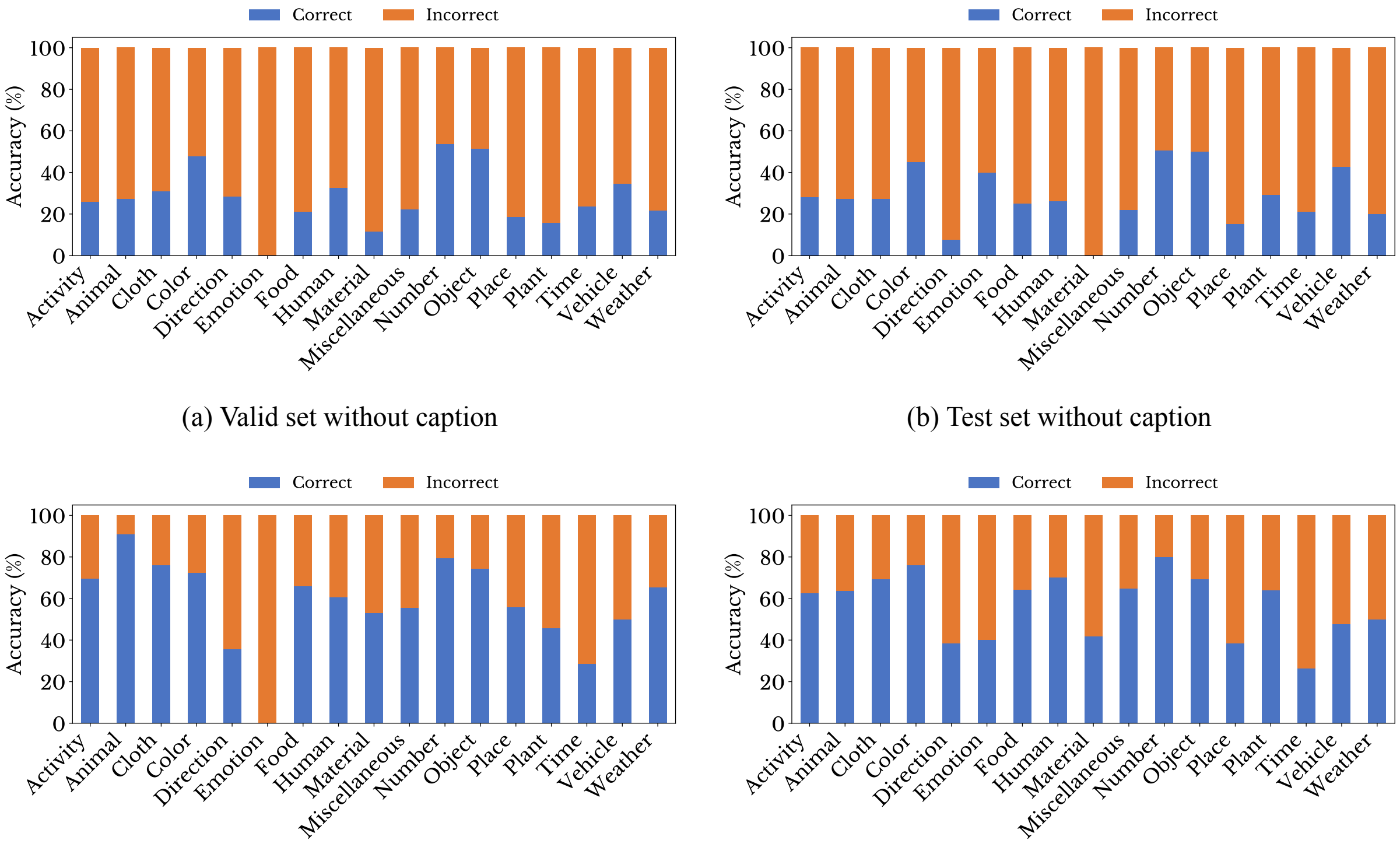

(a) Valid set without caption

(b) Test set without caption

(c) Valid set with caption

(d) Test set with caption

Figure 10: Category-based prediction accuracy by GPT-4 on valid and test sets with and without captions

| | | | |
|---|---|---|---|
| **Image** | 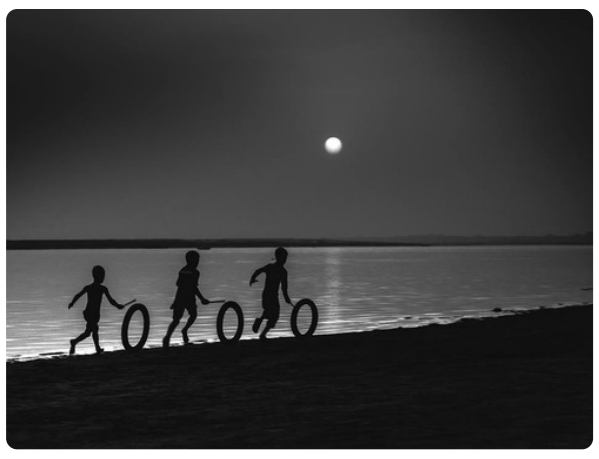 | 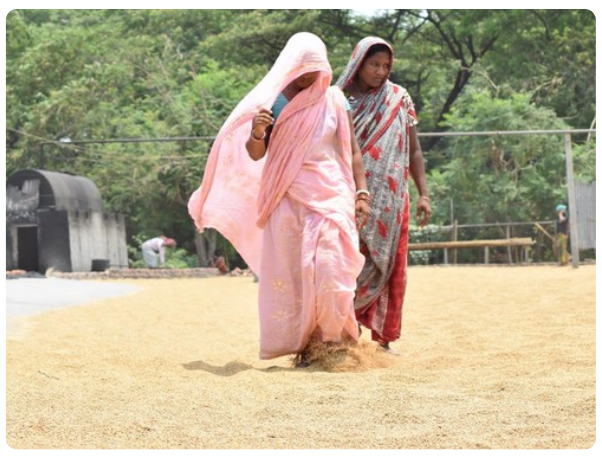 | 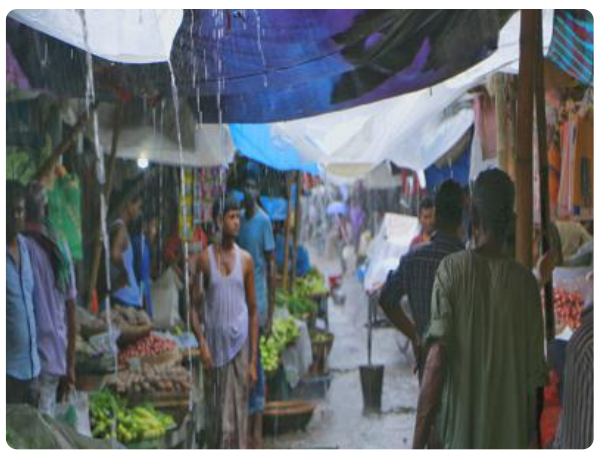 |
| **Caption** | নদীর পাড় দিয়ে তিনজন ছোট ছেলে টায়ার নিয়ে খেলতে খেলতে যাচ্ছে। | শাড়ি পরা দুইজন মহিলা পা দিয়ে ঠেলে ঠেলে ধান শুকাচ্ছে। | বৃষ্টির দিনে একটি কাচা বাজারের দৃশ্য দেখা যাচ্ছে। |
| **Caption *(ENG)*** | Three young boys are playing wih tires by the riverside | Two women wearing saris are drying rice by pushing it with their feet. | On a rainy day, a scene of a local market can be seen |
| **Question** | শিশুরা কি দিয়ে খেলছে? | ছবিতে মহিলারা কী করছে? | ছবিতে দেখা যাচ্ছে কোন ধরনের বাজারের দৃশ্য? |
| **Question *(ENG)*** | What are the kids playing with? | What are the women doing in the picture? | What type of market is seen in the picture? |
| **Answer** | টায়ার (Tire) | ধান শুকাচ্ছে (Drying Paddy) | কাচা বাজার (Fresh Market) |
| **Without Caption** | **GPT-4o:** চাকা<br>**GPT-4o (ENG):** Wheel<br>**GPT-4-turbo:** টায়ার<br>**GPT-4-turbo(ENG):** Tire | **GPT-4o:** চলছে<br>**GPT-4o (ENG):** Walking<br>**GPT-4-turbo:** ধান মাড়াই<br>**GPT-4-turbo(ENG):** Paddy Threshing | **GPT-4o:** সবজি বাজার<br>**GPT-4o (ENG):** Vegetable Market<br>**GPT-4-turbo:** বাজার<br>**GPT-4-turbo(ENG):** Market |
| **With Caption** | **GPT-3.5:** টায়ার<br>**GPT-3.5 (ENG):** Tire<br>**GPT-4o:** টায়ার<br>**GPT-4o (ENG):** Tire<br>**GPT-4-turbo:** টায়ার<br>**GPT-4-turbo(ENG):** Tire | **GPT-3.5:** ধান শুকাচ্ছে। (ধান)<br>**GPT-3.5 (ENG):** Drying Paddy (Paddy)<br>**GPT-4o:** ধান শুকাচ্ছে<br>**GPT-4o (ENG):** Drying Paddy<br>**GPT-4-turbo:** ধান শুকাচ্ছে<br>**GPT-4-turbo(ENG):** Drying Paddy | **GPT-3.5:** কাচা<br>**GPT-3.5 (ENG):** Unripe<br>**GPT-4o:** কাচা বাজার<br>**GPT-4o (ENG):** Fresh Market<br>**GPT-4-turbo:** কাচা বাজার<br>**GPT-4-turbo(ENG):** Fresh Market |

Figure 11: Error Analysis of GPT Models in Our Dataset using with Caption and without Caption Prompting